\documentclass[graybox]{svmult}

\usepackage{type1cm}        
\usepackage{makeidx}         
\usepackage{graphicx}        
\usepackage{multicol}        
\usepackage[bottom]{footmisc}

\usepackage{newtxtext}       %
\usepackage{newtxmath}       

\makeindex             

\begin{document}

\title*{Graph Adapter for Parameter-Efficient Fine-Tuning of EEG Foundation Models}

\author{Toyotaro Suzumura \and\\ Hiroki Kanezashi \and\\ Shotaro Akahori}
\institute{Toyotaro Suzumura, Hiroki Kanezashi \at The University of Tokyo, \email{suzumura@acm.org}, \email{hkanezashi@acm.org}
\and Shotaro Akahori \at Digital Hospital, Inc. \email{akahori@digital-hospital.jp}}
%
%
\maketitle

\abstract{In diagnosing neurological disorders from electroencephalography (EEG) data, foundation models such as Transformers have been employed to capture temporal dynamics. Additionally, Graph Neural Networks (GNNs) are critical for representing the spatial relationships among EEG sensors. However, fine-tuning these large-scale models for both temporal and spatial features can be prohibitively large in computational cost, especially under the limited availability of labeled EEG datasets. We propose EEG-GraphAdapter (EGA), a parameter-efficient fine-tuning (PEFT) approach designed to address these challenges. EGA is integrated into a pre-trained temporal backbone model as a GNN-based module, freezing the backbone and allowing only the adapter to be fine-tuned. This enables the effective acquisition of EEG spatial representations, significantly reducing computational overhead and data requirements. Experimental evaluations on two healthcare-related downstream tasks—Major Depressive Disorder (MDD) and Abnormality Detection (TUAB)—show that EGA improves performance by up to 16.1\% in F1-score compared with the backbone BENDR model, highlighting its potential for scalable and accurate EEG-based predictions.}



\section{Introduction}
\label{sec:1}
Electroencephalography (EEG) is a widely used method for measuring brain activity, capturing electrical signals from various brain regions through electrodes. While other imaging methods, such as functional MRI (fMRI), offer complementary insights into brain function, EEG uniquely provides direct and real-time access to brain dynamics. These signals reveal information related to functions such as consciousness, cognition, and motor activity, with applications spanning the healthcare field, including the diagnosis of mental disorders and the development of brain–computer interfaces (BCIs). 

As such, developing a foundation model for analyzing EEG data across diverse downstream tasks is essential in this domain. In recent years, with the development of large-scale neural network models such as LLMs, many foundation models for time series data have been proposed, including Transformer-based models. Methods for applying these foundation models to EEG signals have also been proposed \cite{bendr,brainbert}.

However, there are several technical challenges in learning EEG representations: 
(1) Prohibitively large computational cost: As the number of parameters for fine-tuning in each downstream task increases, the computational resources for downstream tasks become more significant.
(2) Insufficient training data: The available labeled EEG data sets are limited for many downstream tasks particularly in the healthcare domain due to the considerable effort required for data acquisition, measurement, and labeling by doctors. If fine-tuning is performed with a limited dataset, problems such as overfitting may arise, preventing the model from achieving sufficient accuracy in the downstream task.
In particular, the demand for predictions is increasing in healthcare-related domains that predict neurological disorders and other abnormalities. However, due to issues such as patient privacy, publicly available data is limited.

As for methods to deal with the second problem in downstream tasks, meta-learning \cite{meta-learning} and transfer learning \cite{transfer-learning} have been proposed to leverage knowledge from other similar domains and tasks. They demonstrated the effectiveness of BCI domains, such as motor imagination. However, these approaches require a fine-tuned model trained on similar downstream tasks. Hence, they are hard to apply in cases where there is no sufficiently labeled data or fine-tuned model, such as in disease prediction.

We focus on Parameter-Efficient Fine-Tuning (PEFT), which aims to fine-tune models to achieve representation abilities comparable to fully fine-tuned models, while using fewer computational resources and less data. The main idea of PEFT is to add a lightweight module as an "adapter" with a pre-trained backbone model and fine-tune only the adapter while fixing the backbone model in each downstream task.

Inspired by the concept of PEFT, we propose EEG-GraphAdapter (EGA) to complement the understanding of spatial information in pre-trained models that capture the temporal representations of EEG signals, thereby addressing issues (1) and (2). EGA is a GNN-based module specially designed for downstream tasks by incorporating spatial representations into the input multivariate EEG signals to address the limitations of pre-trained models that only represent the temporal information of EEG.

The BENDR model, pre-trained using stacked convolutional layers on large-scale EEG data, already captures the temporal characteristics of the EEG signals and remains frozen during fine-tuning. This approach allows for efficient learning of spatial features without altering the pre-trained model, enabling effective adaptation to downstream tasks.

EEG signals are typically recorded using electrodes placed at designated locations, and each electrode captures the time-series signals in a different area. By taking into account the spatial arrangement of these electrodes, EEG data can reflect the temporal dynamics and spatial relationships of different brain regions.

Many representation models handle data from individual EEG signals independently, without considering sensor location information or the spatial relationship between sensors.
Spatial EEG information between sensors is essential for downstream tasks such as neurological disorder prediction and emotion recognition. For example, in autism spectrum disorder (ASD), abnormal functional connectivity between the frontal and parietal lobes has been observed \cite{asd1}. Several GNN (Graph Neural Network) models that learn graph structures have been proposed as a model that encodes spatial relationships between EEG sensors for prediction tasks\cite{eeg-gcnn,eeg-gnn,mggcn}.

Integrating GNNs with time-series models such as Transformers makes it possible to capture the temporal dynamics of EEG signals and the spatial relationships between sensors \cite{erdl,spatio-temporal-eeg}. This fusion model addresses the limitations of traditional time-series models in EEG analysis. The main contributions of this paper are as follows.

\begin{itemize}
    \item We propose EEG-GraphAdapter (EGA), a GNN-based module that captures the spatial features between EEG sensors and enhances the representation learning capabilities of the pre-trained BENDR model, while requiring fewer computational resources and less data.
    \item We conducted experiments on two healthcare-related downstream tasks, MDD (Major Depressive Disorder) and Abnormality Detection, demonstrating improvements in both model performance and runtime compared to the baseline BENDR model without the GraphAdapter.
\end{itemize}

\section{Methodology}
\label{sec:2}


\subsection{Model Architecture}

\begin{figure}[ht]
    \centering
    \includegraphics[width=\linewidth]{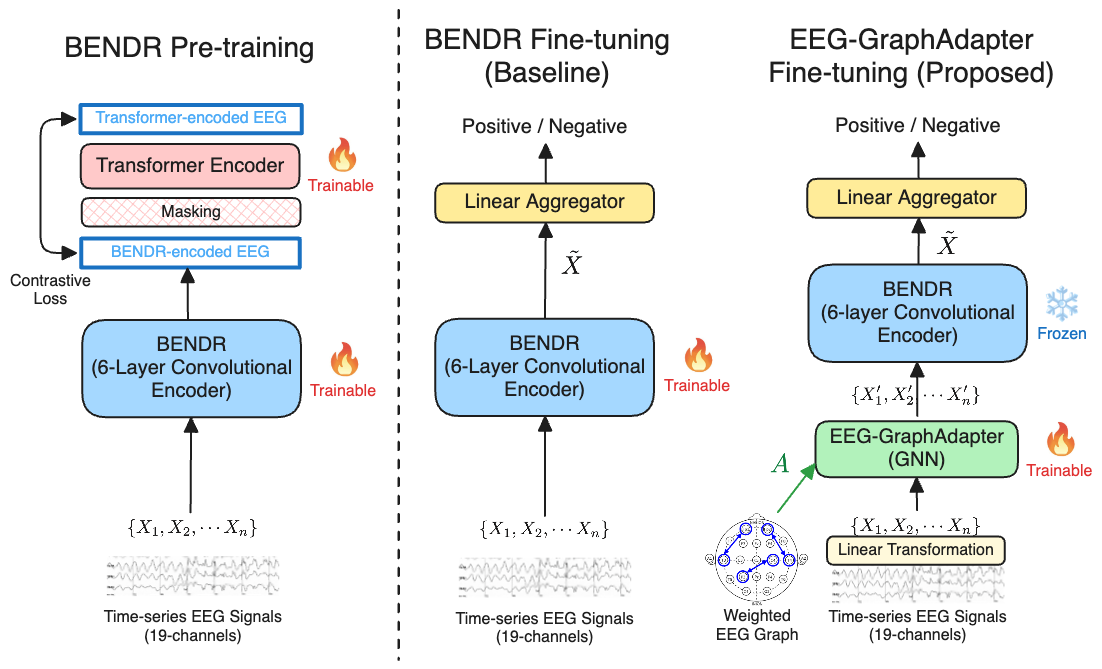}
    \caption{Model Architectures in Pre-training, and Downstream Tasks with Fully Fine-tuned BENDR and Proposed EGA with BENDR Frozen}
    \label{fig:arch}
\end{figure}

Figure \ref{fig:arch} illustrates the architecture for pre-training the time-series BENDR model as the backbone (Left) and for the downstream task using the baseline BENDR model (Center) and the proposed method EGA (Right).

\textbf{BENDR Pre-training} The left subfigure illustrates BENDR's self-supervised pre-training process. The parameters of the BENDR and Transformer models (Transformer Encoder) are updated by a random-masked sequence prediction task, respectively.
The encoded feature vectors are masked, and the model employs self-supervised learning to reconstruct the masked features from the unmasked elements. The masked vectors pass through a Transformer Encoder, which predicts the original unmasked vectors. The model parameters for both the BENDR and the Transformer Encoder are updated using a self-supervised loss function based on the similarity between the predicted and original vectors.

\textbf{BENDR Fine-tuning (Baseline)} The center subfigure shows the baseline approach for downstream fine-tuning, which is consistent with the original BENDR experiments.
We replace the Transformer Encoder with the simpler Linear Aggregator, which better suits classification tasks. Here, the pre-trained BENDR model is fully fine-tuned on the classification task, and its output embeddings are combined into a single feature vector via the Linear Aggregator before the final prediction.

\textbf{EGA Fine-tuning (Proposed)} The right subfigure shows our proposed method, EEG-GraphAdapter (EGA).
In contrast to the baseline approach, the BENDR model is frozen to preserve the representation learned during pre-training. Instead, the EGA module is introduced before the BENDR to incorporate spatial relationships between EEG sensors, improving model performance without fine-tuning the BENDR model.


The following sections describe the BENDR, Linear Aggregator, and EGA used in the proposed method.

\subsubsection{BENDR}
BENDR is a self-supervised model designed to capture the unique characteristics of EEG signals. It comprises six stacked 1D convolutional layers to extract meaningful features from raw EEG signals.
It encodes the original EEG signals $X \in \mathbb{R}^{L\times n}$  into low-dimensional representation feature vectors $\tilde{X} \in \mathbb{R}^{d}$ ($d$ is the dimension of the encoded vector).

BENDR is pre-trained on multivariate EEG signals using self-supervised learning techniques, such as random-masked sequence prediction, to learn robust data representations. The primary objective of BENDR is to facilitate various downstream tasks, even in scenarios with limited computational resources and small datasets. By leveraging self-supervised pre-training, BENDR enables efficient learning and inference, making it well-suited for tasks with constrained data availability.

While Transformer-based models are typically suited for handling time-series data, the computational cost of training on long EEG sequences can be extremely high. To address this, BENDR applies stacked 1D convolutional layers to the raw EEG signals, compressing the sequences into shorter embeddings before passing them into a Transformer model. 
BENDR extends the principles of wav2vec 2.0 \cite{wav2vec2}, which uses a Convolutional Neural Network (CNN) to reduce the dimension of raw audio data before applying a Transformer for pre-training. BENDR adapts this approach to the multi-channel nature of EEG data. 

Although the BENDR model was designed to mitigate the computational cost associated with Transformers on long time-series sequences, when trying to fine-tune a combined model with the BENDR and a GNN in downstream tasks, the number of trainable parameters increases. As a result, additional computational costs and a large amount of data in each downstream task will be required for model convergence.
Furthermore, there is a risk that the representational ability acquired during pre-training on large EEG datasets could be diminished during fine-tuning for specific downstream tasks, a phenomenon known as catastrophic forgetting. 


\subsubsection{Linear Aggregator}
Linear Aggregator aggregates all embeddings encoded by BENDR into a single representation in downstream tasks.
Despite the Transformer Encoder’s integral role in pre-training, experimental results in the original BENDR paper\cite{bendr} show that a simpler Linear Aggregator often yields better performance than the Transformer Encoder when fine-tuning on downstream classification tasks. Consequently, while the Transformer Encoder is necessary during pre-training to learn high-level contextual representations via masked sequence prediction, it is not necessarily optimal to include it in the final classification pipeline. Since the downstream tasks in this study focus on classification based on overall EEG waveform patterns, we consistently utilize the Linear Aggregator.

Each downstream classification task outputs predictions from the representation vectors $\tilde{X} \in \mathbb{R}^{d}$, which are produced by the BENDR model and then passed through the Linear Aggregator and Classifier. First, the Linear Aggregator splits $\tilde{X}$ into four consecutive sub-vectors, and each sub-vector is averaged to produce a fixed-length representation vector $X_{out} \in \mathbb{R}^{4}$ (referred to as the Linear Aggregator). A linear layer with a softmax activation function is applied as a classifier to generate the final predictions.


\subsubsection{EEG-GraphAdapter (EGA)}

To incorporate spatial features into the input features for fine-tuning downstream tasks, we introduce EEG-GraphAdapter (EGA) and place it before the BENDR model.
The objective of the EGA is to predict a label $y \in {y_1, y_2, \ldots}$ from the multivariate EEG signals $X \in \{X_1, X_2, \ldots X_n\} = R^{L \times n}$, where $L$ and $n$ denote the length of the EEG signal and number of EEG sensors (channels), respectively. 
We define a fully connected weighted graph $G = (V, E)$ to represent the relationships between EEG sensors. $V$ denotes the set of EEG sensors ($|V| = n$), and $E \subseteq \{ V \times V \}$ denotes the set of edges between sensors.
We adopt the geodesic distance as the edge weight, using the same approach as EEG-GCNN\cite{eeg-gcnn}.

Depending on the downstream task, EEG sequence samples often have very short sequence lengths. While the BENDR and Transformer models can handle variable-length sequence data as is, the GNN model generally requires the input feature embedding to have a fixed length. For this reason, the raw EEG sequence data is preprocessed to the sequence length required by EGA using a trainable linear layer.

EGA comprises a two-layer GNN model representing spatial relationships between EEG sensors. The GNN processes the temporal EEG signals from each sensor $X$ and generates embeddings $X'$ that incorporate information from other sensors. In this framework, $GNN$ is represented by the EGA, and $A$ is the weighted adjacency matrix.
We explored Graph Convolutional Network (GCN)\cite{gcn}, GraphSAGE\cite{graphsage}, and Graph Attention Network (GAT)\cite{gat} as GNN modules.

\begin{equation}
X' = GNN(X, A)
\end{equation}

The GCN\cite{gcn} model aggregates information by averaging the feature embeddings $X_j$ of adjacent nodes (EEG sensors). In EGA, this approach allows for integrating EEG signal data across sensors, facilitating the learning of sensor relationships.
Because the EEG graph is fully connected, embedding the EEG signal for each sensor aggregates the representation vectors of all other sensors except the sensor itself.

Besides, GraphSAGE\cite{graphsage} aggregates information from neighboring vertices (in this case, all sensors except the target sensor) by randomly sampling from the adjacent vertices. GAT\cite{gat} model adopts an "attention mechanism" that assigns different weights (attention scores) to each neighboring node while aggregating their features. Unlike GCN and GraphSAGE, the GAT model learns individual weights for each node pair, automatically determining which sensor relationships are crucial in predicting labels for each downstream task.

\section{Experiments}
\label{sec:3}

To demonstrate that our proposed EGA can handle classification on downstream tasks using EEG data, we conduct the following experiments to answer the research questions: 

\begin{itemize}
    \item Can EGA, with the integration of the GraphAdapter, improve model performance across various downstream tasks compared to the standalone BENDR model?
    \item Which GNN model is the most effective for the GraphAdapter?
\end{itemize}

\subsection{Experimental Setup}

\subsubsection{BENDR Pre-training}

We adopted the BENDR model as the pre-trained model that forms the backbone of EGA. We pre-trained it from scratch using the latest Temple University EEG Corpus (TUEG) dataset (version 2.0.1, a total of 1.7TB, 69,652 samples) for our experiment. Our proposed EGA can be applied with other pre-trained models handling EEG signals where available.

The TUEG dataset includes EEG data recorded from various devices and patients. We applied preprocessing to make the data suitable for BENDR pre-training. First, we selected only EEG samples that meet the standard 10-20 sensor configuration, excluding samples that did not contain these channels. Additionally, we removed noise, such as artifacts from eye movements and electrical noise from devices. Specifically, we applied 50 Hz and band-pass filters in the 0.1 to 100 Hz range. Then, we standardized the sampling frequency to 256 Hz for all EEG samples.

\subsubsection{Downstream Tasks}

To evaluate the performance of EGA, we conducted binary classification experiments on two healthcare-related downstream tasks (MDD and TUAB), as shown in Table \ref{tb:ds-tasks}.
We used datasets \cite{mdd-data,tuab-data} that are publicly available and have explicit labels for abnormal (positive) and healthy (negative) samples for downstream tasks.

\begin{description}
    \item[MDD]\cite{mdd-data} This task predicts patients with Major Depressive Disorder (MDD; positive) and healthy control subjects (negative) based on EEG data.
    \item[TUAB]\cite{tuab-data} It distinguishes whether the EEG sample is abnormal (positive) with sustained spikes or patterns such as Periodic Lateralized Epileptiform Discharges (PLEDs) or Generalized Periodic Epileptiform Discharges (GPEDs), or normal (negative) with the Posterior Dominant Rhythm (PDR) typically appears when a subject is relaxed with their eyes closed.
\end{description}

The MDD dataset \cite{mdd-data} contains 19-channel EEG signals from patients with Major Depressive Disorder (MDD: positive) and healthy control (negative) subjects. For each subject, there are EEG samples measured during rest with eyes open (EO), eyes closed (EC), and a task in which the subject responds to letters displayed on the screen. In this experiment, we split the samples from EO and EC signals into 60-second segments and used them as sample data for evaluation.
TUAB \cite{tuab-data} involves predicting whether each EEG sample from The TUH Abnormal EEG Corpus is labeled normal (negative) or abnormal (positive). Given the imbalance in this dataset, we randomly selected 20 patients for each label and split each patient's EEG sequence into 60-second segments for evaluation.

\begin{table}[ht]
    \centering
    \begin{tabular}{l|l|r|r|r|r|r|r}
    \hline
    Name & Task & Length (s) & Freq(Hz) & \#Subjects & \#Samples & \%Positive & Folds \\ \hline
    MDD & Major Depressive Disorder & 60 & 256 & 63 & 126 & 49 & 10 \\
    TUAB & Abnormality Detection & 60 & 250 & 40 & 517 & 49 & 5 \\ \hline
    \end{tabular}
    \caption{Datasets for Downstream Tasks}
    \label{tb:ds-tasks}
\end{table}

We applied the following preprocessing steps to the downstream task datasets to ensure consistency with the TUEG dataset used in the BENDR model's pre-training. First, all EEG data were re-sampled to a 256 Hz sampling frequency. Additionally, although some datasets contain EEG sensor data beyond the standard 19 channels used in the 10-20 protocol, these extra channels were excluded.

Similar to the preprocessing in BENDR pre-training, we applied a 50 Hz notch filter and a 0.1-100 Hz bandpass filter to remove noise from measurement devices and other artifacts. Lastly, for EEG samples—except those from the TUAB dataset—with sequence lengths shorter than 60 seconds (256 Hz $\times$ 60 s = 15,360 samples), we applied a linear transformation to adjust sequence length using a linear layer.

\subsubsection{Hardware and Software Configurations}

Our experiments were conducted on an instance of our academic cloud platform named "mdx"\cite{mdx}, equipped with a single NVIDIA A100 GPU (40GB), two Intel Xeon Platinum 8368 CPUs (each with 38 cores at 2.4 GHz), and 512 GiB of DRAM.

We implement EGA as the extension of BENDR, which is implemented in \footnote{\url{https://github.com/SPOClab-ca/BENDR}} with PyTorch backend. We also implement GNN models (GCN, GraphSAGE, and GAT) as EGA modules in PyTorch Geometric.
These GNN models have two hidden layers of size 64, and the number of attention heads was set to one for the GAT model.
We used Adam for the optimizer, with a fixed learning rate of 0.00001.

For each downstream task, the dataset was split according to k-fold cross-validation (k=10 for MDD and k=5 for TUAB), followed by fine-tuning over 7 epochs.

\subsection{Performance of EGA}




To validate the effectiveness of the EGA, we compared the results of each downstream task against the baseline BENDR model. The architectures of the baseline and the proposed EGA are illustrated in Figure \ref{fig:arch}. In this configuration, the input includes time-series EEG data from 19 channels and a fully connected weighted EEG graph, where the distance between channels determines the weights. We adopt graph convolutional network (GCN), GraphSAGE, and Graph Attention Network (GAT) models as the EEG-GraphAdapter.

Tables \ref{tb:mdd-result} and \ref{tb:tuab-result} compare models on the MDD and TUAB downstream tasks, respectively. The baseline performance, achieved through fine-tuning the BENDR model, is compared with the EGA utilizing GCN, GraphSAGE, and GAT (EGA-GCN, EGA-GraphSAGE, EGA-GAT, respectively).
Since these downstream tasks have nearly balanced positive and negative samples, we compare the model performance with two metrics: F1-score and AUROC. The number in bold indicates the best performance for each metric, and the underlined values indicate the better performance than the baseline.

In the MDD downstream task, the GAT-based EEG-GraphAdapter (EGA-GAT) achieved a 12.8\% higher F1-score and a 2.7\% higher AUROC than the baseline BENDR model, respectively. Meanwhile, the GCN-based EEG-GraphAdapter (EGA-GCN) also outperformed the baseline, indicating that both GAT and GCN can effectively capture spatial information for MDD. However, the performance of EGA-GraphSAGE fell below the baseline, particularly in the F1-score.

On the other hand, in the TUAB downstream task, EGA-GraphSAGE achieved a 16.1\% improvement over the baseline in the F1-score. The GraphSAGE-based model consistently outperformed the baseline across all metrics, demonstrating stable and high model performance.
Overall, these results suggest that the most suitable GNN architecture can vary depending on the downstream task, underscoring the importance of selecting an appropriate graph-based module for each application.

\begin{table}[ht]
    \centering
    \begin{tabular}{c|c|c}
    Model & AUROC & F1-score \\ \hline
    Baseline & 0.9407 & 0.7581 \\ \hline
    EGA-GCN & \underline{0.9576} & \underline{0.8065} \\
    EGA-GraphSAGE & 0.8305 & 0.4032 \\
    EGA-GAT & \textbf{0.9661} & \textbf{0.8548} \\
    \end{tabular}
    \caption{Model Performance in MDD Downstream Task}
    \label{tb:mdd-result}
\end{table}

\begin{table}[ht]
    \centering
    \begin{tabular}{c|c|c}
    Model & AUROC & F1-score \\ \hline
    Baseline & 0.7377 & 0.4328 \\ \hline
    EGA-GCN & 0.7126 & \underline{0.4387} \\
    EGA-GraphSAGE & \textbf{0.7459} & \textbf{0.5027} \\
    EGA-GAT & 0.6983 & 0.3958 \\
    \end{tabular}
    \caption{Model Performance in TUAB Downstream Task}
    \label{tb:tuab-result}
\end{table}

\subsection{Effectiveness of EGA}

During downstream tasks, the BENDR model is kept frozen, allowing only the parameters of the EGA to be updated. This approach aims to efficiently learn the relationships between EEG sensors with minimal computational cost. It is anticipated that if the BENDR model is sufficiently pre-trained, further training of the BENDR model during downstream tasks may not be necessary. On the other hand, fine-tuning the BENDR model during these tasks makes it possible to capture time-series information more accurately for each task, although the computational cost increases.

The number of trainable parameters for each model (BENDR and three EGA models with two hidden layers with size 64) is shown in Table \ref{tb:num-params}. Compared to fine-tuning the BENDR model with many parameters, fine-tuning only EGA is expected to complete the task in roughly 1/6 to 1/4 of the time. The number of parameters in GraphSAGE is approximately twice that of GCN and GAT due to the PyG implementation, where the parameter matrices for linear transformations are implemented separately for self-nodes and neighbor nodes in PyTorch Geometric.

\begin{table}[ht]
    \centering
    \begin{tabular}{c|r}
    Model & \#Params \\ \hline
    BENDR & 6,459,257 \\
    EGA-GCN & 998,432 \\
    EGA-GraphSAGE\ & 1,981,472 \\
    EGA-GAT & 1,029,216 \\ \hline
    \end{tabular}
    \caption{Number of Trainable Parameters}
    \label{tb:num-params}
\end{table}

Theoretically, training time should be reduced in proportion to the number of model parameters, but the speedup was a maximum of about 17\%. This is due to overheads other than model parameter updates, such as pre-fetching of EEG datasets. By parallelizing the data loading process, this performance bottleneck should be mitigated, but this is beyond the scope of our research.

\section{Related Work}

\subsection{EEG and Transformer-based Models}

Similar to the BENDR\cite{bendr} model, numerous methods have been proposed to pre-train Transformer-based models on EEG signals using self-supervised learning to enhance few-shot learning performance in downstream tasks. For instance, BrainWave \cite{brainwave} introduces a Transformer-based foundation model that can be pre-trained on both EEG signals and more invasive but highly accurate intracranial electroencephalograph (iEEG) signals.
MAEEG \cite{maeeg} also performs pre-training by masking and reconstructing embeddings derived from stacked convolutional layers, similar to BENDR. However, MAEEG uses a reconstruction loss instead of a contrastive loss, achieving more stable and accurate results on longer EEG signals than BENDR.
EEG2Rep \cite{eeg2rep} applies multiple masking techniques, such as semantic subsequence preserving (SSP), which consecutively masks semantically related parts of the data. This allows the model to learn more abstract features during pre-training, resulting in a robust pre-trained model resistant to noise in raw EEG data.

\subsection{GNN Models in Spatial Representation}

In addition to Transformer-based models, several GNN-based models have been proposed to account for the relationships between EEG sensors. MGGCN \cite{mggcn} constructs a Brain Functional Connectivity Network (BFCN) based on the functional connections between sensors derived from EEG signals and uses a GCN-based model to predict Major Depressive Disorder (MDD). Similarly, EEG-GCNN \cite{eeg-gcnn} represents the relationships between EEG sensors as a fully-connected graph with weights based on sensor distances and the correlations of time-series data using a GCN-based model for neurological disorder prediction.
EEG-GNN \cite{eeg-gnn} takes a similar approach by representing the sensor relationships as a weighted graph, with weights based on sensor distances. This model, using GraphSAGE or GIN, predicts subjects' motor patterns.

These GNN-based models represent the relationships between EEG sensors as a graph structure, learning to capture functional connectivity between different brain regions. However, except for EEG-GNN, the time-series EEG signals corresponding to graph vertices (sensors) are typically simplified to focus on frequency-domain intensity. This approach does not account for the specific short- and long-term signal variations that Transformer-based models are designed to capture.

\subsection{PEFT and Graph Adapter}
Many methods have been proposed for PEFT that are designed to reduce the number of parameters. One typical example is Low-Rank Adaptation (LoRA)\cite{lora}, which reduces the computation costs for fine-tuning by adding low-rank matrices to the top of the pre-trained model.
In addition, several methods have been proposed to combine an LLM with a GNN as an adapter. This makes integrating the LLM's semantic representation with the graph's spatial information possible.

For example, G-Adapter \cite{g-adapter} enhances molecular structure prediction by inserting parameters into the feed-forward network (FFN) of a pre-trained Transformer model, allowing for incorporating spatial information, such as adjacency matrices, during fine-tuning. Recently, PEFT techniques have gained traction in applying large-scale pre-trained models like LLMs to various applications. Specifically, methods have been proposed that combine LLMs and GNNs for tasks involving Text-Attributed Graphs (TAGs), where text information is attached to nodes in the graph, such as in Knowledge Graphs.

One such method, TAPE \cite{tape}, encodes textual features such as article titles in citation networks using a pre-trained LLM and fine-tunes only a small-scale language model and a GNN. This approach maintains the LLM's semantic richness while improving performance on graph-structured classification tasks. Similarly, GraphAdapter \cite{graphadapter} fixes the parameters of a pre-trained LLM during downstream tasks, updating only the GNN and a fusion layer that integrates both models, thus improving efficiency on prediction tasks in TAGs, such as citation networks and social networks.

In this study, we apply the PEFT concept to downstream tasks involving EEG data and propose EGA, which captures both the temporal dynamics and the relationships between EEG sensors. We leverage the BENDR model, pre-trained on general time-series EEG data, allowing task-specific adaptation for each downstream task through the GNN-based adapter.

\section{Conclusion and Future Work}

In this study, we proposed EEG-GraphAdapter (EGA), a method designed to efficiently solve healthcare-related downstream tasks with EEG signals by introducing a GNN model that learns the relationships between EEG sensors. 
With a GNN-based module as an adapter, we efficiently utilized pre-trained models, such as BENDR, that capture the time-series representation of EEG signals. This allowed for efficient learning of EEG representations tailored to each downstream task with minimal computational cost.
Through experiments on the two downstream tasks, EGA demonstrated both superior accuracy and shorter training times compared to the baseline that fully fine-tuned BENDR. These results highlight the effectiveness of EGA in achieving high performance with reduced computational requirements.

As for future work, we aim to evaluate the broader applicability of the proposed EGA across various pre-trained models and downstream tasks. While this study employed the convolution-based BENDR model as the pre-trained model, we will explore other Transformer-based models such as BrainWave (Brant-2)\cite{brainwave}. Given that Transformer-based models typically require substantial computational resources for fine-tuning, the graph adapter may prove even more effective in such contexts.
Additionally, we plan to evaluate and verify the performance of EGA in other healthcare-related downstream tasks, where the relationship between EEG sensors is considered to have a significant impact on the prediction results.
For instance, in autism spectrum disorder (ASD), weak functional connectivity between the frontal and temporal lobes or between the cerebral hemispheres has been observed. In contrast, abnormalities in functional connectivity between the prefrontal cortex and the striatum have been observed in attention deficit hyperactivity disorder (ADHD).
Since the amount of EEG data sets related to neurological disorders is often limited, the effect of introducing EGA could be particularly pronounced in these cases. Therefore, EGA offers a promising approach for resource-limited clinical EEG applications, paving the way for more accurate and scalable neurological disorder detection.

\begin{acknowledgement}
This work is partially supported by JSPS KAKENHI Grant JP21K17749 and JP23K28098.In this research work, we used the “mdx: a platform for building data-empowered society”.
\end{acknowledgement}

\end{document}